\documentclass{bmvc2k}

\newlength\savedwidth
\newcommand{\whline}{\noalign{\global\savedwidth\arrayrulewidth
                            \global\arrayrulewidth 1.5pt}%
                   \hline
                   \noalign{\global\arrayrulewidth\savedwidth}}

\newcommand{\swhline}{\noalign{\global\savedwidth\arrayrulewidth
\global\arrayrulewidth 1.2pt}%
\hline
\noalign{\global\arrayrulewidth\savedwidth}}

\usepackage[floatrowsep=none]{floatrow}
\newfloatcommand{capbtabbox}{table}[][\FBwidth]
\usepackage{bm}
\usepackage{enumitem}
\usepackage{floatrow}
\usepackage{multirow}
\usepackage{subfigure}
\usepackage[ruled]{algorithm2e}
\usepackage{graphicx}
\usepackage{amsmath,amssymb} 
\usepackage{tabularx}
\usepackage{colortbl}
\definecolor{Gray}{gray}{0.9}

\title{Highly Efficient Regression for Scalable Person Re-Identification}

\addauthor{Hanxiao Wang}{hanxiao.wang@qmul.ac.uk}{1}
\addauthor{Shaogang Gong}{s.gong@qmul.ac.uk}{1}
\addauthor{Tao Xiang}{t.xiang@qmul.ac.uk}{1}

\addinstitution{
 Vision Group,\\
 School of Electronic Engineering and Computer Science,\\
 Queen Mary, University of London\\
 London, E1 4NS, UK
}

\runninghead{Wang et al.}{Highly Efficient Regression for Scalable Person Re-id}

\begin{document}

\maketitle

\begin{abstract}
Existing person re-identification models are poor for scaling up to
large data required in real-world applications due to: (1) Complexity:
They employ complex models for optimal performance resulting in high
computational cost for training at a large scale; (2) Inadaptability: Once trained,
they are unsuitable for incremental update to incorporate any new data
available. This work proposes a truly scalable solution to re-id by
addressing both problems.  Specifically, a Highly Efficient Regression
(HER) model is formulated by embedding the Fisher's criterion to a
ridge regression model for very fast re-id model learning with
scalable memory/storage usage. Importantly, this new HER model
supports faster than real-time incremental model updates therefore
making real-time active learning feasible in re-id with
human-in-the-loop.  Extensive experiments show that such a simple and fast
model not only outperforms notably the state-of-the-art re-id methods,
but also is more scalable to large data with additional benefits to
active learning for reducing human labelling effort in re-id
deployment.

\end{abstract}

\section{Introduction}
\label{sec:intro}
Person re-identification~(re-id) refers to the problem of
matching people across disjoint camera views
at different locations and times~\cite{gong2014re}.
It is challenging since
a person's appearance often undergoes dramatic changes in a multi-camera environment, affected by viewing angles, body poses, illuminations and background clutters.
Re-id is further compounded by the homogeneous clothing styles among different individuals.
As a result,
the concentration of most existing state-of-the-art methods~\cite{KISSME_CVPR12,
PRD_PAMI13,CVPR13LFDA,Zhao_MidLevel_2014a,wang2014person,wang2016pami,li2013learning,
Li_DeepReID_2014b,xiong2014person,liao2015person,Liao_2015_ICCV,Ahmed2015CVPR,zhang2016learning}
has been anchored at reducing intra-person appearance disparity
whilst enhancing inter-person appearance individuality
to optimise person identity-discriminative information.
By deploying such learning
principles on exhaustively pre-labelled pairwise training data,
these methods have reported ever-increasing
accuracies on existing re-id benchmarks.



Then, should we expect from them a scalable re-id solution for deployment in a
real-world environment?
The answer is no.
Specifically, a truly
scalable re-id system~\cite{REIDchallenge} requires:
(1) Low model complexity
with scalable computational cost and memory usage in model training;
and (2) High model adaptability
supporting fast model update to incorporate
any new and increasingly larger data, e.g. unknown camera pairs.
None of the aforementioned state-of-the-art
satisfies these two requirements.
By fixating on the re-id matching performance only, existing re-id models are typically
solved by either slow iterative optimisation~\cite{li2013learning,
Li_DeepReID_2014b,Zhao_MidLevel_2014a,PRD_PAMI13,Liao_2015_ICCV,Ahmed2015CVPR}
or costly generalised eigenproblems~\cite{CVPR13LFDA,xiong2014person,liao2015person,zhang2016learning},
requiring hours or even days to train when
the data size grows over $10^4$.
Moreover, most of the models are restricted to learning in a batch-mode, therefore failing
to offer a solution to
efficient model update without re-training from scratch when new labelled data become available.
In a real-world application scenario, a human operator using a re-id
system will generate new labels in the process of deployment. It is
highly desirable that a re-id model can grow continuously its knowledge whilst being used.
Given the existing re-id models, this can only be achieved by
re-training from scratch, putting
a huge burden on both model updating time and data
storage\footnote{Need to keep all the data for training.}, rendering
existing models poorly suited for scaling up to cumulatively large data.
Moreover, such re-training cannot provide immediate
responses to the operator, which is
a fundamental flaw for any systems with human-in-the-loop.

\begin{figure}[!t]
\centering
\includegraphics[width=0.93\textwidth]{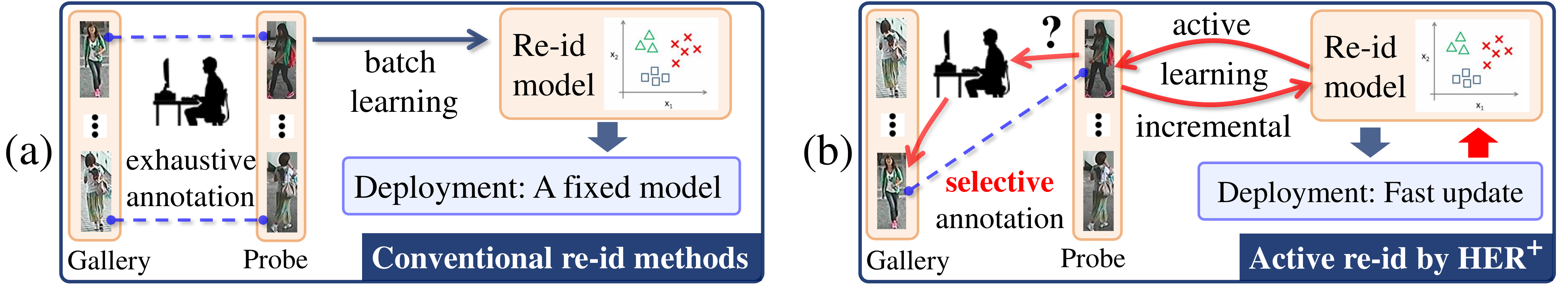}
\vspace{-0.4cm}
\caption{\small (a) Conventional re-id: A re-id model is trained
on a fully labelled training set, then fixed for deployment;
(b) Active re-id by HER$^{+}$: A training set is actively labelled
incrementally on-the-fly as a re-id model is incrementally learned,
and further updated without re-training during future deployment.}
\label{fig:framework}
\end{figure}

To make person re-id more scalable for real-world deployment, a model
must be simple with very fast learning and inference algorithms and
supports incremental learning. To that end, we propose an extremely
simple but very fast re-id model by exploring multivariate ridge
regression~\cite{hastie2005elements} to enable regression capable of
performing re-id verification tasks.  The new model, Highly Efficient
Regression~(HER) for re-id, has four advantages over existing methods:
(1) Low model complexity with a very simple and fast closed-form
solution, involved with only a set of linear equations. It is readily scalable to
large data if more efficient algorithms like LSQR~\cite{paige1982lsqr} are used.
(2) By embedding Fisher's
criterion~\cite{fisher1936use,fukunaga2013introduction}, HER does not
sacrifice any discriminative power for speed, and outperforms the
state-of-the-art re-id models with complex algorithms.
(3) HER$^{+}$,
an additional incremental formulation of HER, enables real-time incremental model
update to incorporate new data of which the existing batch-based re-id
models are incapable.
(4) HER$^{+}$ makes {\em active learning re-id} with
human-in-the-loop (Fig.~\ref{fig:framework}) feasible to reduce
human labelling costs. This is important as heavy human supervision is
another bottleneck to scalable re-id.
%
%

{\bf Contributions}: (1) A Highly Efficient Regression (HER)
re-id model is formulated. The model is both very simple, can be
implemented by one line of code, and scalable to large data.  (2) An
incremental extension HER$^{+}$ is further formulated for real-time
model update to incorporate new labelled data.  (3) HER$^{+}$ in
active learning is explored with three novel joint
exploration-exploitation criteria proposed for active re-id.
Extensive experiments (Sec.~\ref{sec:experiment}) on three benchmark
datasets VIPeR~\cite{VIPeR}, CUHK01~\cite{CUHK_dataset} and
CUHK03~\cite{Li_DeepReID_2014b} have shown the superiority of the
proposed HER model over seven state-of-the-art models, including
Mid-level Filter~\cite{Zhao_MidLevel_2014a},
Deep+~\cite{Ahmed2015CVPR}, kLFDA~\cite{xiong2014person},
XQDA~\cite{liao2015person}, MLAPG~\cite{Liao_2015_ICCV},
NFST~\cite{zhang2016learning}, and Ensembles~\cite{Anton_2015_CoRR}, on both
re-id performance and efficiency. Our experiments also
reveal the benefits of HER$^{+}$ for active learning in re-id with
reduced human annotation effort.

\vspace{-0.3cm}
\section{Related Work}
\vspace{-0.3cm}
\label{sec:related}
Most existing state-of-the-art in
re-id~\cite{PCCA_CVPR12,KISSME_CVPR12,PRD_PAMI13,CVPR13LFDA,Zhao_MidLevel_2014a,
  xiong2014person,wang2014person,wang2016pami,Li_DeepReID_2014b,ding2015deep,
  Anton_2015_CoRR,liao2015person,Liao_2015_ICCV} focus only on
learning the person identity-discriminative information for pursuing
high performance.  Existing approaches include exploring
view-invariant visual
features~\cite{SDALF_CVPR10,SalienceReId_CVPR13,FisherVectorReId_ECCV12,Stan14ColorName},
imposing pairwise or list-wise constraints for
ranking~\cite{Anton_2015_CoRR,chen2015relevance,wang2016pami,loy2013person},
discriminative subspace/distance metric learning~\cite{KISSME_CVPR12,
  PRD_PAMI13,CVPR13LFDA,xiong2014person,liao2015person,Liao_2015_ICCV,zhang2016learning},
and deep
learning~\cite{ShiZLLYL15,ding2015deep,Li_DeepReID_2014b,Ahmed2015CVPR}.
However, existing models have poor scalability due to two factors:
High complexity to train; and low adaptability to update.  For
example, many methods rely on iterative 
optimisation\cite{li2013learning,
  Li_DeepReID_2014b,Zhao_MidLevel_2014a,PRD_PAMI13,Liao_2015_ICCV,Ahmed2015CVPR}
which are extremely slow in training. The current fastest closed-form
solutions are those solved by generalised
eigenproblems~\cite{CVPR13LFDA,xiong2014person,liao2015person,zhang2016learning}
which are still expensive to compute.  Moreover, most contemporaries
only assume a batch-mode learning scheme: To incorporate any new data, a
system has to  keep all the past training data and re-train a new model
from scratch. This re-training approach makes them unscalable to
large-scale deployment in the real-world.

Ridge regression~\cite{hoerl1970ridge,hastie2005elements}, as a
regularised least squares model, is one of the most well-studied
machine learning models. It has a simple closed-form solution solved
by a linear system, and thus low model complexity.  Furthermore, many
well-optimised algorithms~\cite{paige1982lsqr} can be readily
applied to large data.  Finally, its adaptable solution
supports efficient model update for incremental
learning~\cite{liu2009least}. The new HER model casts re-id into such a
regression problem, benefiting from all of its advantages in
scalability.  To explore ridge regression for discriminative re-id
verification tasks, the proposed HER model is further embedded with
the criterion of Fisher Discriminant
Analysis (FDA)~\cite{fisher1936use,fukunaga2013introduction} to encode
person identity-discriminative information.  The
relationship between FDA and linear regression has been studied
for binary~\cite{duda2012pattern} and
multi-class~\cite{hastie2005elements,park2005relationship}
classification tasks.  Recently,
similar connections have been discovered for their regularised
counterparts~\cite{zhang2009flexible,
  zhang2010regularized,lee2015equivalence}. However, this work is the
first to formulate it for a verification setting as in re-id.  For its
incremental extension, our model HER$^{+}$, differs significantly to
~\cite{liu2009least} which only supports updates on a single sample
without regularisation employed.


The proposed HER$^{+}$ enables active learning in re-id.
Active learning~\cite{kang2004using,osugi2005balancing,cebron2009active,hospedales2012unifying,ebert2012ralf,loy2012stream}
argues that not all samples are equally important to
model learning, and thus requires the training set to be
actively selected on-the-fly as the model
is updated with new labels.
For active learning to be plausible in engaging human-in-the-loop, real-time
model response with a very fast updating algorithm is essential.
A recent attempt at active re-id was
reported in~\cite{das2015active}. Instead of learning a generalised
cross-view matching function, it trains multi-class SVM
person classifiers on known identities. The model cannot be deployed
to re-identify a person without having abundant person-specific training images in advance.
Thus, it is not applicable to most existing benchmarks~\cite{VIPeR,CUHK_dataset,Li_DeepReID_2014b}
nor common re-id settings where
each training person is often only captured by a
single or a very few shot(s) and during testing the person to be re-identified
does not assume to have any training data~\cite{REIDchallenge}. Moreover, their
model updates still require expensive re-training.
Compared to~\cite{das2015active}, the proposed HER$^{+}$ with
active learning criteria learns a
generalised re-id matching function, specially designed for the most common re-id conditions.
Moreover, it achieves better
human effort efficiency, since HER$^{+}$ can be
incrementally updated whenever new labels arrive,
and in turn immediately benefits further active
data selection process in a loop.

\vspace{-0.6cm}
\section{Methodology}
\vspace{-0.3cm}
In the conventional re-id setting,
a pre-labelled training set consists of $n$
cross-view images
and their identity labels are assumed to exist,
denoted as $\{\mathbf{X} \in \mathbb{R}^{d\times n}, \mathbf{l}\in \mathbb{Z}^{n}\}$,
where $\mathbf{X} = [\bm{x}_1,\cdots,\bm{x}_n]$,
$\bm{x}_i \in \mathbb{R}^d$ is the $i$-th image's feature vector,
$l_i \in \{1,\cdots,c\}$ is its identity label, 
and $c$ is the total amount of person identities.
After model training, the learned generalised re-id model is deployed to match person images of
some independent probe and gallery population
by measuring their inferred distances.


\subsection{Highly Efficient Regression for Re-Id}
\label{sec:HER}
Given labelled data $\{\mathbf{X} \in \mathbb{R}^{d\times n}, \mathbf{l}\in \mathbb{Z}^{n}\}$,
we aim to learn a discriminative projection $\mathbf{P} \in \mathbb{R}^{d\times k}$.
%
Specifically, we expect that in the projected $k$-dim subspace,
the samples of the same person become closer,
and those of different people are further apart.

Suppose there exists a set of {\it `ideal'} vectors $\bm{y}_i$ in the $\mathbb{R}^k$ subspace
which perfectly satisfy the above discriminative relationship
to represent each training sample $\bm{x}_i$,
and together they construct a matrix $\mathbf{Y}\in\mathbb{R}^{n\times k}$ in a row-wise manner.
Treating $\mathbf{Y}$ as regression outputs and $\mathbf{P}$ the regression parameter,
a discriminative projection can thus be estimated with multivariate ridge regression,
which minimises a least mean squared error
between $\mathbf{Y}$ and the projected features of training samples, $\mathbf{X}^\top \mathbf{P}$,
with the magnitude of $\mathbf{P}$ being $l_2$-regularised:
\begin{equation}
\resizebox{.40\hsize}{!}{$
\mathbf{P} = \arg\min_\mathbf{P} \; \frac{1}{2}\|\mathbf{X}^\top\mathbf{P} - \mathbf{Y}\|_F^2 + \lambda\|\mathbf{P}\|_F^2,
$}
\label{eq:HER_obj}
\end{equation}
where
$\|\cdot\|_F$ is the Frobenius norm,
$\lambda$ controls the regularisation strength. $\mathbf{Y}$ is
known as an indicator matrix.
The above formulation leads to a simple closed-form solution:
\vskip -10pt
\begin{equation}
\resizebox{.23\hsize}{!}{$
\mathbf{P} = \big(\mathbf{X}\mathbf{X}^\top + \lambda\mathbf{I}\big)^\dagger\mathbf{X}\mathbf{Y},
$}
\label{eq:HER_solution}
\end{equation}
where $(\cdot)^\dagger$ denotes a Moore-Penrose inverse, and $\mathbf{I}$ an identity matrix.

To frame the re-id task as a regression problem with
Eq.~\eqref{eq:HER_solution}, we need to find the optimal regression outputs, $\mathbf{Y}$,
so that the person identity-discriminative information
is encoded by the learned projection.
To tackle this problem, we propose to 
explore the criterion of
FDA~\cite{fisher1936use,fukunaga2013introduction}.
Specifically, in FDA the intra-person appearance disparity and
inter-person appearance difference are measured by two scatter matrices:
\begin{equation}
\resizebox{.8\hsize}{!}{$
\mathbf{S}_w = \frac{1}{n}\sum_{j=1}^c \sum_{l_i=j} (\bm{x}_i - \bm{u}_j)(\bm{x}_i - \bm{u}_j)^\top, \qquad
\mathbf{S}_b = \sum_{j=1}^c n_j(\bm{u}_j - \bm{u})(\bm{u}_j - \bm{u})^\top,
$}
\end{equation}
where $\mathbf{S}_w$ and $\mathbf{S}_b$ are denoted as within-class and between-class
scatter respectively, $\bm{u}_j$ and $\bm{u}$ being the class-wise means and the global mean,
and $n_j$ the sample size of class $j$.
Then a discriminative assignment of  $\mathbf{Y}$ for re-id can be
obtained by optimising the FDA's criterion,
with the earlier derived representation in Eq.~\eqref{eq:HER_solution} as a constraint:
\begin{equation}
\resizebox{.7\hsize}{!}{$
\mathbf{Y} = \arg\max_{\mathbf{Y}} \; trace\Big(\big(\mathbf{P}^\top\mathbf{S}_w\mathbf{P}\big)^\dagger\big(\mathbf{P}^\top\mathbf{S}_b\mathbf{P}\big)\Big), \quad
s.t. \; \mathbf{P} = \big(\mathbf{X}\mathbf{X}^\top + \lambda\mathbf{I}\big)^\dagger\mathbf{X}\mathbf{Y}.
$}
\label{eq:FDA_HER}
\end{equation}
There exists a simple solution~\cite{liu2009least,ye2007least} $\mathbf{Y}\in \mathbb{R}^{n\times k}$
to Eq.~\eqref{eq:FDA_HER} as follows:
\begin{equation}
\resizebox{.6\hsize}{!}{$
k = c; \quad
\mathbf{Y}_{ij} = \frac{1}{\sqrt{n_j}}, \;\;\; \text{if} \;\; l_i=j; \quad \text{and} \quad
\mathbf{Y}_{ij} = 0, \;\;\; \text{if} \;\; l_i\neq j.
$}
\label{eq:Y}
\end{equation}
Equation~\eqref{eq:Y} has three interesting properties:
(1) Output matrix $\mathbf{Y}$ can be directly constructed
with negligible cost before training.
Once $\mathbf{Y}$ is set,
our Highly Efficient Regression (HER) solution for re-id
can be directly acquired by Eq.~\eqref{eq:HER_solution};
(2) Unlike existing re-id approaches~\cite{CVPR13LFDA,xiong2014person,liao2015person,zhang2016learning}
which also exploit Fisher's criterion, the
calculations of $\mathbf{S}_w$ and $\mathbf{S}_b$ are no longer
required by HER's solution in Eq.~\eqref{eq:HER_solution}.
(3) By setting $\mathbf{Y}$ as Eq.~\eqref{eq:Y},
all samples of the same identity tend to be projected onto one single point,
thus the learned subspace becomes maximally
person identity-discriminative.
A similar idea was considered by a recent study~\cite{zhang2016learning}
which learns a discriminative null space for projection.
But since no regularisation is used in \cite{zhang2016learning},
the learned null space  on training data is often not optimal for
testing due to over-fitting. In contrast,
the HER model is less prone to over-fitting because of regularisation
(Eq.~\eqref{eq:HER_obj}), and achieves better re-id performance
(Sec.~\ref{sec:experiment}). 

On computational efficiency,
HER is much more efficient than
iterative optimisation re-id approaches, e.g.~\cite{li2013learning,
Li_DeepReID_2014b,Zhao_MidLevel_2014a,PRD_PAMI13,Liao_2015_ICCV,Ahmed2015CVPR},
since our closed-form solution is
significantly simpler and faster in computation;
Compared to those models based on solving eigenproblems~\cite{xiong2014person,    
CVPR13LFDA,liao2015person,zhang2016learning},
HER's solution is also more efficient.
For a feature dimension $d$,
sample size $n$, and $m = \min(d,n)$,
performing an eigen-decomposition to solve
a generalised eigenproblem or a null space
requires $\frac{3}{2}dnm + \frac{9}{2}m^3$ floating point addition/multiplications~\cite{penrose1955generalized},
whereas solving the linear system in Eq.~\eqref{eq:HER_solution}
takes $\frac{1}{2}dnm + \frac{1}{6}m^3$~\cite{cai2008srda}.
This gives maximally 9 times speed-up by HER (see
Sec.~\ref{sec:experiment} for empirical results).

As a linear model, HER can be further
kernalised to capture non-linear transforms.
Assume the kernel matrix is $\mathbf{K}\in \mathbb{R}^{n\times n}$,
then kernelised projection vectors $\mathbf{Q} \in \mathbb{R}^{n\times c}$
are:
\begin{equation}
\resizebox{.22\hsize}{!}{$
\mathbf{Q} = \big( \mathbf{K}\mathbf{K}^\top + \lambda\mathbf{K} \big)^\dagger\mathbf{K}\mathbf{Y}.
$}
\label{eq:ker}
\end{equation}

\subsection{Incremental HER$^{+}$}
\label{sec:incHER}
Most conventional re-id models, and the proposed HER,
are still limited in their scalability,
since they only consider a batch learning scheme:
To update a model with newly labelled data,
they have to add the new data to the overall training pool
and re-train from scratch.
In a real-world where data cumulation can increase significantly,
such a re-training strategy will become extremely expensive,
e.g. take hours or days to perform even one update.

To further improve the scalability of HER, we introduce an incremental
learning formulation HER$^{+}$,
enabling fast model updates without the need for re-training from scratch.
%
Suppose at time $t$,
$\mathbf{X}_t \in \mathbb{R}^{d\times n_t}$ is
the features of $n_t$ previously labelled images
of $c_t$ person identities,
$\mathbf{Y}_t \in \mathbb{R}^{n_t \times c_t}$ is their indicator matrix defined by Eq.~\eqref{eq:Y};
$\mathbf{X}' \in \mathbb{R}^{d\times n'}$ the features of $n'$ newly labelled
images of $c'$ new person classes.
and $\mathbf{Y}' \in \mathbb{R}^{n' \times (c_t+c')}$
is its corresponding indicator matrix defined by Eq.~\eqref{eq:Y}.
Let $\mathbf{T}_t = \mathbf{X}_t\mathbf{X}_t^\top + \lambda\mathbf{I}$,
then HER's projection $\mathbf{P}_t \in \mathbb{R}^{d \times c_t}$ at time $t$
can then be written as $\mathbf{P}_{t}   = \mathbf{T}_{t}^\dagger \mathbf{X}_{t} \mathbf{Y}_{t}$
(Eq.~\eqref{eq:HER_solution}). Next, we derive an incremental update
version of HER with a two-step updating scheme.

\noindent \textbf{Updating $\mathbf{T}^\dagger$ - }
The first step is to update matrix $\mathbf{T}^\dagger$. After incorporating the new data at $t$,
the updated data matrix and indicator matrix can be represented as:
\begin{equation}
\resizebox{.38\hsize}{!}{$
\mathbf{X}_{t+1} = [\mathbf{X}_t, \; \mathbf{X}'],
\quad
\mathbf{Y}_{t+1} =
\Big[ \begin{array}{c}
\mathbf{Y}_t\oplus\mathbf{O}\\
\mathbf{Y}'  \end{array} \Big],
$}
\label{eq:XY}
\end{equation}
where we define operator $(\cdot)\oplus\mathbf{O}$
as augmenting a matrix by padding appropriate numbers of zero columns on the right.
Since $\mathbf{T}_t = \mathbf{X}_t\mathbf{X}_t^\top + \lambda\mathbf{I}$,
we write $\mathbf{T}_{t+1}$ as:
\begin{equation}
\resizebox{.2\hsize}{!}{$
\mathbf{T}_{t+1} = \mathbf{T}_t + \mathbf{X}'\mathbf{X}'^{\top}.
$}
\label{eq:T}
\end{equation}
Applying the Sherman-Morrison-Woodbury formula~\cite{woodbury1950inverting} to Eq.~\eqref{eq:T},
the update of $\mathbf{T}^\dagger$ is:
\begin{equation}
\resizebox{.42\hsize}{!}{$
\mathbf{T}_{t+1}^{\dagger} = \mathbf{T}_t^\dagger - \mathbf{T}_t^\dagger
\mathbf{X}'\big(\mathbf{I}+\mathbf{X}'^{\top}\mathbf{T}_t^\dagger\mathbf{X}'\big)^\dagger
\mathbf{X}'^{\top}\mathbf{T}_t^\dagger.
\label{eq:Tdagger}
$}
\end{equation}

\noindent {\bf Updating $\mathbf{P}$ - }
Next we update projection matrix $\mathbf{P}$.
Eq.~\eqref{eq:HER_solution} and Eq.~\eqref{eq:XY} together give:
\begin{equation}
\resizebox{.54\hsize}{!}{$
\mathbf{P}_{t+1} = \mathbf{T}_{t+1}^\dagger \mathbf{X}_{t+1} \mathbf{Y}_{t+1}
 = (\mathbf{T}_{t+1}^\dagger \mathbf{X}_t \mathbf{Y}_t)\oplus\mathbf{O}
 + \mathbf{T}_{t+1}^\dagger \mathbf{X}'\mathbf{Y}'.
 $}
\end{equation}
Expanding $\mathbf{T}_{t+1}^\dagger$ in the first term with Eq.~\eqref{eq:Tdagger}, and
consider $\mathbf{P}_{t}   = \mathbf{T}_{t}^\dagger \mathbf{X}_{t} \mathbf{Y}_{t}$,
the update of $\mathbf{P}$ is:
\begin{equation}
\resizebox{.58\hsize}{!}{$
\mathbf{P}_{t+1}  = \Big(\mathbf{P}_t - \mathbf{T}_t^\dagger
\mathbf{X}'\big(\mathbf{I}+\mathbf{X}'^{\top}\mathbf{T}_t^\dagger\mathbf{X}'\big)^\dagger
\mathbf{X}'^{\top}\mathbf{P}_t\Big)\oplus\mathbf{O}
 + \mathbf{T}_{t+1}^\dagger \mathbf{X}'\mathbf{Y}' .
 $}
 \label{eq:update}
\end{equation}
The above updating scheme (Eq.~\eqref{eq:Tdagger}
and Eq.~\eqref{eq:update}) forms our HER$^{+}$ algorithm.
It shows that the previous data $\{\mathbf{X}_t,\mathbf{Y}_t\}$
is not needed for model updates.
Analogous to Eq.~\eqref{eq:ker},
kernelisation can also be applied as a pre-processing step.

\noindent \textbf{Implementation consideration - }
The HER$^{+}$ algorithm
supports updates on both a single and/or a small chunk of data
with $n'\geqslant 1$. If the data chunk size $n'\ll d$, where $d$ is
the feature dimension, it is faster to 
perform $n'$ separate updates on each new sample
instead of by chunk. The reason is that
in such a way
the Moore-Penrose matrix inverse in Eq.~\eqref{eq:Tdagger}
and Eq.~\eqref{eq:update} can be reduced to $n'$ separate
scaler inverses whose computation is much cheaper.

\subsection{Active Re-Id by Joint Exploration-Exploitation}
\label{sec:active}
The efficient model updates achieved by HER$^{+}$
makes active learning plausible in a re-id system
in order to reduce human labelling effort.
To differentiate from the aforementioned conventional setting,
we define a more scalable {\it active re-id} setting as follows:

\noindent \textbf{Active re-id - }An {\it unlabelled} probe set $\widetilde{\mathcal{P}}$
and gallery set $\widetilde{\mathcal{G}}$  are the only available data before training.
Assume at time step $t \in \{1,\cdots,\tau\}$ where $\tau$ is
a limited human labelling budget, $m_t$ denotes the currently learned re-id model,
$\widetilde{\mathcal{P}_t}$ and $\widetilde{\mathcal{G}_t}$
denote the remaining unlabelled data at time $t$.
{\it Active re-id} describes the following procedure:
(1) Image(s) $\mathcal{I}_t^p \in \widetilde{\mathcal{P}_t}$ of a new training probe identity $l_t$
 is actively selected by $m_t$, according to its usefulness/importance measured by
certain active sampling criteria;
(2) A ranking list of unlabelled gallery images $\widetilde{\mathcal{G}}_t$
against the selected probe is then generated by $m_t$;
(3) Human annotators verify $\mathcal{I}_t^p$'s cross-view matching image(s) $\mathcal{I}_t^g \in \widetilde{\mathcal{G}}_t$ in the ranking list;
(4) Model $m_{t+1}$ is updated from new
annotation $(\mathcal{I}_t^p,\mathcal{I}_t^g,l_t)$, and repeat from step~(1).

To select the samples that would maximise model discrimination
capacity, we propose a joint exploration-exploitation
active sampling strategy, consists of three criteria (Fig.~\ref{fig:active}):

\begin{figure}[t]
\centering
\includegraphics[width=0.95\textwidth]{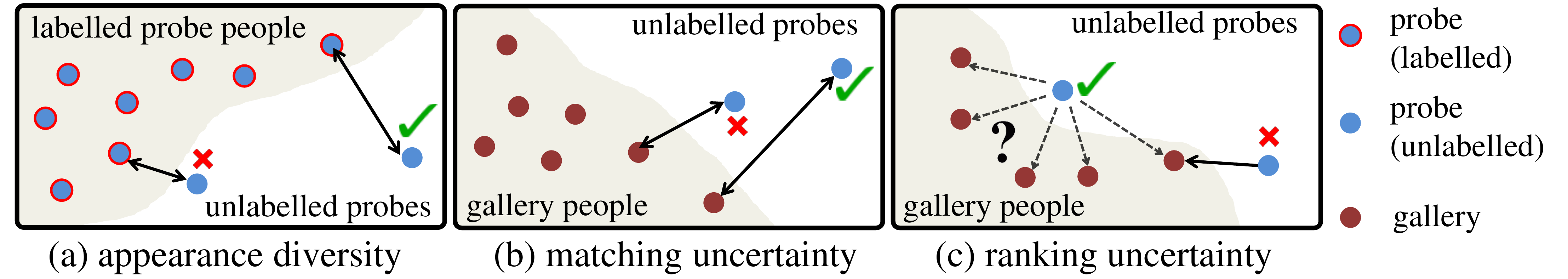}
\vskip -0.4cm
\caption{Joint exploration-exploitation criteria for active re-id.}
\label{fig:active}
\end{figure}

\noindent {\bf Appearance diversity exploration - }
The diversity of training persons'
appearances is critical for
a re-id model to generalise well, thus the
preferred next probe image to annotate should
lie in the most unexplored part within the population.
Assume at time $t$,
the distance between any two
samples $(\bm{x}_1,\bm{x}_2)$ obtained by the current re-id model $m_t = \mathbf{P}_t$ is:
\begin{equation}
d(\bm{x}_1,\bm{x}_2|m_t) = (\bm{x}_1 - \bm{x}_2)^\top\mathbf{P}_t\mathbf{P}^\top_t(\bm{x}_1 - \bm{x}_2).
\label{eq:dist}
\end{equation}
Let $\widetilde{\mathcal{P}}_t$
and $\mathcal{P}_t$
denote the unlabelled and labelled part of probe set $\widetilde{\mathcal{P}}$ at time $t$
respectively ($\widetilde{\mathcal{P}}_t \bigcup \mathcal{P}_t = \widetilde{\mathcal{P}}$), we thus measure the diversity degree of
an unlabelled probe sample ${\bm{x}}_i^p \in \widetilde{\mathcal{P}}_t$ by the
distance to its {\it within-view nearest neighbour} in $\mathcal{P}_t$ (Fig.~\ref{fig:active} (a)):
\begin{equation}
\varepsilon_1({\bm{x}}_i^p)
= \min \; d({\bm{x}}_i^p, \bm{x}_j^p|m_t), \quad s.t. \;\; {\bm{x}}_i^p \in \widetilde{\mathcal{P}}_t, \;\; \bm{x}_j^p \in \mathcal{P}_t.
\end{equation}

\noindent {\bf Matching uncertainty exploitation - }
Uncertainty-based exploitative sampling schemes has been widely
researched for classification tasks~\cite{joshi2009multi,settles2008analysis,ebert2012ralf},
querying the least certain sample for human to annotate.
Tailor-made for re-id tasks, our second criterion here
prefers the probe samples staying far away from the gallery after projection at time $t$,
i.e. the re-id model $m_t$ remains unclear on what are the corresponding
cross-view appearances of these `poorly-matched' probe images.
We measure the matching uncertainty of an unlabelled probe sample
${\bm{x}}_i^p \in \widetilde{\mathcal{P}}_t$ by the distance to its
{\it cross-view nearest neighbour} in 
$\widetilde{\mathcal{G}}$ (Fig.~\ref{fig:active} (b)):
\begin{equation}
\varepsilon_2({\bm{x}}_i^p)  = \min\; d({\bm{x}}_i^p, \bm{x}_j^g|m_t), \quad s.t. \;\; {\bm{x}}_i^p \in \widetilde{\mathcal{P}}_t, \;\; \bm{x}_j^g \in \widetilde{\mathcal{G}}.
\end{equation}


\noindent \textbf{Ranking uncertainty exploitation - }
Due to similar appearances among different identities,
a weak re-id model could generate close ranking scores for those
visually-ambiguous gallery identities to a given probe.
It would thus be useful to ask human to label such a probe sample
to enhance a re-id model's discrimination power (Fig.~\ref{fig:active} (c)).
We define a distribution over all gallery samples
$\bm{x}^g_j \in \widetilde{\mathcal{G}}$, conditioned on a given probe $\bm{x}^p_i$ according
to their ranking scores:
\begin{equation}
p_{m_t}(\bm{x}^g_j|\bm{x}^p_i) = \frac{1}{Z_{i}^t}{e^{-d({\bm{x}}_i^p, \bm{x}_j^g|m_t)}}, \quad \text{where} \;\; Z_{i}^t = {\sum_k e^{-d({\bm{x}}_i^p, \bm{x}_k^g|m_t)}}, \;\; \bm{x}_k^g \in \widetilde{\mathcal{G}}.
\end{equation}
Large entropy of the above distribution means that the ranking
scores are close to each other, indicating $m_t$'s uncertainty on its
returned ranking list. Thus, our third criterion is:
\begin{equation}
\varepsilon_3({\bm{x}}_i^p) = - \sum_j p_{m_t}(\bm{x}^g_j|\bm{x}^p_i) \log p_{m_t}(\bm{x}^g_j|\bm{x}^p_i), \quad s.t. \;\; {\bm{x}}_i^p \in \widetilde{\mathcal{P}}_t, \;\; \bm{x}_j^g \in \widetilde{\mathcal{G}}.
\end{equation}

\noindent {\bf Joint exploration-exploitation - }
Similar to \cite{cebron2009active,ebert2012ralf},
we combine exploitation and exploration into our final active selection strategy.
Specifically, our final selection criterion is a 
sum of $\varepsilon_1, \varepsilon_2, \varepsilon_3$, where each score is normalised to $(0,1)$:
\vspace{-.1cm}
\begin{equation}
\label{eqn:final_active}
\varepsilon(\bm{x}^p_i) = \varepsilon_1(\bm{x}^p_i) +
 \varepsilon_2(\bm{x}^p_i) + \varepsilon_3(\bm{x}^p_i).
\end{equation}
The unlabelled probe samples
can thus be sorted according to this selection criterion,
and the one with highest score is then
selected for human annotation.
%
Finally, whenever such newly labelled data is obtained, our HER$^{+}$
model can perform an immediate incremental update.
Deployed together with the above active sampling criteria,
our HER$^{+}$ model helps to focus human effort only on labelling
those most useful samples, thus maximises the cost-effectiveness
in learning a scalable re-id model.


%

\vspace{-0.5cm}
\section{Experiments}
\label{sec:experiment}
\vspace{-0.2cm}
We conducted experiments under both the conventional supervised re-id setting
and the new active re-id setting to evaluate our
HER~(Sec.~\ref{sec:HER}) and HER$^{+}$~(Sec.~\ref{sec:incHER}) models
respectively. We also evaluated the proposed active sampling
criteria~(Sec.~\ref{sec:active}).

\noindent \textbf{Datasets and features -}
Three benchmarks,
VIPeR~\cite{VIPeR}, CUHK01~\cite{CUHK_dataset}, and CUHK03~(manual version)~\cite{Li_DeepReID_2014b}
were considered for our evaluation.
VIPeR contains a total of 632 people with one image per person per view,
CUHK01 contains 971 people with two images per person per view,
whilst CUHK03 contains 13,164 images of 1,360 people with a
maximum of five images per person per view. By default the Local
Maximal Occurrence (LOMO) features~\cite{liao2015person} (26,960
dimensions) are adopted for image representation.

\vspace{-0.5cm}
\subsection{Conventional Re-Id Evaluation}
\vspace{-0.2cm}
\noindent \textbf{Settings and Comparisons - } We first considered the standard
fully-supervised re-id setting to evaluate the proposed HER
model~(Sec.~\ref{sec:HER}). Seven state-of-the-art baselines were compared:
Mid-level Filter~\cite{Zhao_MidLevel_2014a}, Deep+~\cite{Ahmed2015CVPR},
kLFDA~\cite{xiong2014person}, XQDA~\cite{liao2015person}, MLAPG~\cite{Liao_2015_ICCV},
NFST~\cite{zhang2016learning}, and
Ensembles~\cite{Anton_2015_CoRR}. Whenever code is available and
features can be replaced, we compared them using the same LOMO features.
In addition, we also compared with the Ensembles fusion model in which
four types of features with two different matching metrics are
fused~\cite{Anton_2015_CoRR}. 
For comparison, we fused two types of
features for HER: LOMO~\cite{liao2015person} and that of
~\cite{KCCAReid}. For each feature type we trained one HER model
independently and performed a final score-level fusion (HER (fusion)).
For data partitions, VIPeR and CUHK01 were
randomly split into two equal halves for training/testing
as in~\cite{liao2015person,zhang2016learning},
and repeated by 10 times for averaging. On CUHK03, we adopted the
standard 20 training/testing splits~\cite{Li_DeepReID_2014b} and the
single-shot testing protocol~\cite{liao2015person,zhang2016learning}. In implementation,
we applied the same RBF kernel settings of kLFDA~\cite{xiong2014person} and NFST~\cite{zhang2016learning}.
The free parameter $\lambda$ in Eq.~\eqref{eq:HER_obj}
and the parameters of other models were all determined by
cross-validation.

\begin{table}[t]
\centering
\scalebox{0.76}{
\renewcommand{\arraystretch}{1.05}
\setlength{\tabcolsep}{0.26cm}
\begin{tabular}{l|cccc|cccc|cccc}
\whline
dataset  & \multicolumn{4}{c|}{VIPeR~\cite{VIPeR}} & \multicolumn{4}{c|}{CUHK01~\cite{CUHK_dataset}} & \multicolumn{4}{c}{CUHK03~\cite{Li_DeepReID_2014b}} \\ 
rank     & R1   & R5   & R10   & R20  & R1   & R5   & R10   & R20   & R1   & R5   & R10   & R20   \\ \hline \hline
$l_2$ norm&15.6 & 27.7 & 36.2  & 49.2 & 20.5 & 37.1 & 45.3  & 55.3  & 11.3 & 27.3 & 39.8  &  55.8 \\ 
Mid-level~\cite{Zhao_MidLevel_2014a}& 29.1 & 52.3 & 66.0  & 79.9 & 34.3 & 55.1 & 65.0  & 74.9  & - & - & - & -  \\ 
Deep+~\cite{Ahmed2015CVPR}& 34.8 & 63.6 & 75.6  & 84.5 & 47.5 & 71.6 & 80.3  & 87.5  & 54.7 & 86.5 & 93.9  & 98.1  \\ 
kLFDA~\cite{xiong2014person}& 38.6 & 69.2 & 80.4  & 89.2 & 54.6 & 80.5 & 86.9  & 92.0  & 45.8  & 77.1 & 86.8  & 93.1  \\ 
XQDA~\cite{liao2015person}& 40.0 & 68.1 & 80.5  & 91.1 & 63.2 & 83.9 & 90.0  & 94.2  & 52.2 & 82.2 & 92.1  & 96.3  \\ 
MLAPG~\cite{Liao_2015_ICCV}& 40.7 & 69.9 & 82.3  & 92.4 & 64.2 & 85.4 & 90.8  & 94.9  & 58.0 &  \bf 87.1 & 94.7  & \bf  98.0  \\ 
NFST~\cite{zhang2016learning}& 42.3 & 71.5 & 82.9  & 92.1 & 65.0 & 85.0 & 89.9  & 94.4  & 58.9 & 85.6 & 92.5  & 96.3  \\ 
\rowcolor{Gray} \bf HER~(LOMO) & \bf 45.1 & \bf 74.6 & \bf 85.1  & \bf 93.3 & \bf 68.3 & \bf 86.7 & \bf 92.6 & \bf 96.2 & \bf 60.8 & 87.0 & \bf 95.2  & 97.7  \\ \hline \hline
Ensembles~\cite{Anton_2015_CoRR}& 45.9 & 77.5 & 88.9  & \bf 95.8 & 53.4 & 76.4 & 84.4 & 90.5  & 62.1 & 89.1 & 94.3 & 97.8 \\ 
\rowcolor{Gray} \bf HER~(fusion) & \bf 53.0 & \bf 79.8 & \bf 89.6  & 95.5 & \bf 71.2 & \bf 90.0 & \bf 94.4 & \bf 97.3 & \bf 65.2 & \bf 92.2 & \bf 96.8  & \bf 99.1  \\ \whline
\end{tabular}}
\vskip -8pt
\caption{Re-id comparisons under the conventional setting ($100\%$
  labelled training set).}
\label{tab:CMC_conventional}
\end{table}

\noindent \textbf{Results - } The Cumulated Matching Characteristics
(CMC) curve is adopted as the evaluation metric.
The results are shown in Table~\ref{tab:CMC_conventional}.
It is evident that HER
significantly outperforms all existing competitors on all three datasets.
Taking the Rank-1 recognition rate as a comparison,
HER~(LOMO) has notably improved the current state-of-the-art NFST~\cite{zhang2016learning} from
$42.3\%$ to $45.1\%$ on VIPeR, from $65.0\%$ to $68.3\%$ on CUHK01,
and from $58.9\%$ to $60.8\%$ on CUHK03 when the same LOMO feature is used.
This shows the superiority of HER's regularised discriminative
projection over the null-space model NFST.
Although both models encourage the same class samples to be projected onto a single
point, the NFST model does not employ regularisation therefore
is sensitive to over-fitting.
Through fusing different feature representations, HER model's
performance can be further boosted.
Specifically, HER~(fusion) outperforms the Ensembles fusion
model~\cite{Anton_2015_CoRR}, with Rank-1 rates of
$53.0\%$, $71.2\%$ and $65.2\%$ for the three benchmarks respectively,
$7.1\%$, $17.8\%$, and $3.1\%$ better than the Ensembles;
despite that the Ensembles model fuses four feature types and two re-id matching models, whilst a single
HER model fuses only two feature types.
%
\vspace{-0cm}
\begin{figure} [b]
\begin{floatrow}
\hskip -15pt
\capbtabbox{
\scalebox{0.75}{
	\renewcommand{\arraystretch}{1}
	\setlength{\tabcolsep}{0.3cm}
\begin{tabular}{l|c|c|c|c}
\swhline
time(sec)& \bf HER & kLFDA & XQDA    & MLAPG   \\ \hline
VIPeR   &  \bf 1.2& 5.0  & 4.1    & 50.9           \\
CUHK01  &\bf 4.2 & 45.9 & 51.9   & 746.6          \\
CUHK03  &  \bf 248.8&  2203.2 &  3416.0   &   4.0$\times 10^4$       \\ \whline
\end{tabular}
}}{%
\vspace{-10pt}
\caption{\small Training time for conventional setting.}
\label{table:time_conv}
}
\hskip -8pt
\capbtabbox{
\scalebox{0.8}{
	\renewcommand{\arraystretch}{1.26}
	\setlength{\tabcolsep}{0.22cm}
\begin{tabular}{l|c|c|c}
\swhline
$\text{time}_U$(sec) & VIPeR   & CUHK01 & CUHK03  \\ \hline
\bf HER$^{+}$   &  \bf 0.03  & \bf 0.2 &  \bf 0.6   \\
HER & 0.4  & 2.3 & 53.8 \\
\swhline
\end{tabular}
}}
{%
\vspace{-10pt}
\caption{\small Model updating time~(50\% labelled).}
\label{table:time_act}
}
\end{floatrow}
\end{figure}

To evaluate the computational efficiency of HER on model training time, critical
for scaling up to large data, we chose three
representative existing models with code available: kLFDA, XQDA, and MLAPG.
Whilst kLFDA and XQDA are solved by generalised eigenproblems (closed-form),
MLAPG is computed by iterative optimisation using
line-searching. We recorded each model's training time in
each trial and averaged over 10 trials. Our experiments
were performed on a Linux server@2.6GHz CPU with 384GB memory.
Table~\ref{table:time_conv} reports the averaged results: HER takes
$1.2$ seconds to train on a small dataset VIPeR~(over $10^2$
images). This is respectively $\times4.2$ times, $\times3.4$ times,
and $\times42.4$ times faster than kLFDA, XQDA, and MLAPG.
On a much larger dataset CUHK03~(over $10^4$ images),
comparing to kLFDA (2,203.2 sec), XQDA (3,416.0 sec) and MLAPG
(40,000.0 sec), HER only takes 248.8 seconds to train. That is
respectively $\times8.8$ times, $\times13.7$ times, and $\times 160.8$
times faster. This shows significant advantage of HER over other
models for scaling up to larger data. Although we could not evalute the
training time of Deep+ model~\cite{Ahmed2015CVPR}
due to unavailable code, typical deep model training time is
expected much longer than those in Table~\ref{table:time_conv}.

\subsection{Active Re-Id Evaluation}
\noindent \textbf{Settings - } Next we
consider the active re-id setting~(Sec.~\ref{sec:active})
to evaluate our HER$^{+}$ algorithm and active sampling criteria.
Our experiments were conducted on VIPeR and CUHK01.
For each dataset, we train HER$^{+}$ model incrementally,
with the next probe sample randomly sampled~(Random),
or actively selected by the new joint exploration-exploitation
criteria~(Joint$\text{E}^2$). We also compared with
another active sampling strategy~\cite{ebert2012ralf}
which finds the densest region of the probe sample
space~(Density).
The groundtruth matched-images of selected probes were given as simulated
  human-in-the-loop labelling.
As in a realistic re-id system, an operator
can only afford to actively label a small proportion of
the vast amount of unlabelled data, we varied
the labelling budget from $10\%$ to $50\%$ of the overall training set.
Our experiments were conducted in 10 trials with the averaged results reported.

\begin{table}[t]
\centering
\scalebox{0.65}{
	\renewcommand{\arraystretch}{1.05}
	\setlength{\tabcolsep}{0.3cm}
\begin{tabular}{c|l|cccccc|cccccc}
\whline
& dataset  & \multicolumn{6}{c|}{VIPeR~\cite{VIPeR}} & \multicolumn{6}{c}{CUHK01~\cite{CUHK_dataset}}  \\ 
rank &labelled ($\%$)    & 10   & 20   & 30   & 40 & 50 & 100 & 10   & 20   & 30   & 40 & 50 & 100\\ \hline \hline
\multirow{ 3}{*}{R1(\%)} & Random  & 15.8 & 20.8 & 27.2 & 30.1  & 35.2 & \bf 45.2 & 21.4 & 34.4 & 44.8  & 48.4 & 53.7 & \bf 64.7\\
& Density~\cite{ebert2012ralf} & 15.7 & 23.3 & 27.6 & 30.4  & 34.2 & \bf 45.2 & 23.5 & 36.4 & 46.3  & 49.1 & 53.9 & \bf 64.7\\
& \cellcolor{Gray}{\bf Joint$\mathbf{E}^2$} & \cellcolor{Gray}\bf 18.2  & \cellcolor{Gray}\bf 25.7 & \cellcolor{Gray}\bf 30.7 & \cellcolor{Gray}\bf 34.6 & \cellcolor{Gray}\bf 37.1 &\cellcolor{Gray}\bf 45.2 &  \cellcolor{Gray}\bf 29.9 & \cellcolor{Gray}\bf 41.0 & \cellcolor{Gray}\bf 47.2 &  \cellcolor{Gray}\bf 51.6  & \cellcolor{Gray}\bf 55.8 & \cellcolor{Gray}\bf 64.7 \\ \hline \hline

\multirow{ 3}{*}{R5(\%)} &Random & 35.8 & 45.2 & 51.8 & 57.1  & 61.4 & \bf 74.5& 44.1 & 58.6  & 68.5  & 73.5 & 77.1&\bf 85.2\\
& Density~\cite{ebert2012ralf} & 38.1 & 50.7 & 57.1 & 59.4  & 64.2 & \bf 74.5 & 47.4 & 63.4 & 70.4  & \bf 75.2 & 78.2 & \bf 85.2\\
&\cellcolor{Gray}\bf Joint$\mathbf{E}^2$ & \cellcolor{Gray}\bf 41.0 &  \cellcolor{Gray}\bf 51.0 & \cellcolor{Gray}\bf 57.2 &\cellcolor{Gray}\bf 61.7  & \cellcolor{Gray}\bf 66.8 &\cellcolor{Gray}\bf 74.5 & \cellcolor{Gray}\bf 54.9  & \cellcolor{Gray}\bf 64.8 & \cellcolor{Gray}\bf 71.1 &  \cellcolor{Gray} 75.1  & \cellcolor{Gray}\bf 79.8& \cellcolor{Gray}\bf 85.2\\ \whline

\end{tabular}}
\vskip -8pt
\caption{\small Re-id performances under active re-id scheme  (10\% to 50\% training data labelled). We also include
incrementally training with 100\% (fully) labelled training data for comparison to Table~\ref{tab:CMC_conventional}.}
\label{tab:CMC_active}
\end{table}

\noindent \textbf{Results - } Table~\ref{tab:CMC_active}
shows the   Rank-1 and Rank-5 recognition rates on both datasets.
It is evident that:
(1)
On VIPeR, the Rank-1 rate of Joint$\text{E}^2$
is averagely $3.4\%$ higher than Random, and it also compares favourably against the
alternative Density criteria~\cite{ebert2012ralf}.
(2)
Active re-id achieves better performances with less human labelling efforts.
For example, on VIPeR active labelling only $30\%$ of the data
achieves a $30.7\%$ Rank-1 rate, already higher than that of
randomly labelling $40\%$ of the data.
Also, when applied with Joint$\text{E}^2$, only trained with
$50\%$ labelled data on CUHK01, the HER$^{+}$ model achieved $55.8\%$
for Rank-1, already higher than the Mid-level, Deep+, and kLFDA
models trained on $100\%$ fully labelled training set (Table~\ref{tab:CMC_conventional}).
(3)
When incrementally trained with $100\%$ labelled training set,
HER$^{+}$ achieved comparable results with
the batch-based HER (Table~\ref{tab:CMC_conventional}),
suggesting that incremental updates of HER$^{+}$ does not
sacrifice re-id performances.

Finally we evaluated computational costs of incremental model update
versus re-training.
A further experiment was conducted:
We first trained a HER model
by using $50\%$ labelled training data on each dataset.
We then provided additional one pair of
matched images as newly labelled data from online confirmation, and updated the trained HER model as
follows:
(1) Incremental update by HER$^+$ on the two newly arrived images;
(2) Re-train HER using an enlarged training data pool consisting of the 
initial 50\% labelled data plus the new pair.
The model updating times
are presented in Table~\ref{table:time_act} (averaged results over 10 trials).
It is evident that
HER$^{+}$ only takes 0.03 second, 0.2 second and 0.6 second
respectively to perform real-time model update on the three
benchmarks. This is $\times 13.3$ times, $\times 11.5$ times
and $\times 89.7$ times faster than re-training HER in batch mode.
In particular, on a larger dataset CUHK03, the advantage of incremental model
update by HER$^{+}$ over re-training is more significant.
Given that HER is already the fastest batch learning model
over others by 1-2 orders of magnitude
(Table~\ref{table:time_conv}), the incremental model update by
HER$^{+}$ is thus over $\times 10^4$ times faster
than re-training a conventional existing model such as kLFDA, XQDA and
MLAPG. This offers significant advantage in scaling up to large data
in real-world deployments.

\section{Conclusion}
\label{sec:conclusion}
A highly scalable re-id model HER is proposed by formulating re-id as a
ridge regression problem. Compared to existing re-id models,
HER is much faster and cheaper to compute, whilst yields superior re-id performance.
An incremental HER$^{+}$ model is introduced to enable
sustainable incremental cumulative model update and to facilitate active learning
re-id for reducing human supervision in re-id deployment.
Extensive experiments show the scalability of HER and HER$^{+}$
and their potential for large size data re-id deployments.

\bibliography{reid}
\end{document}